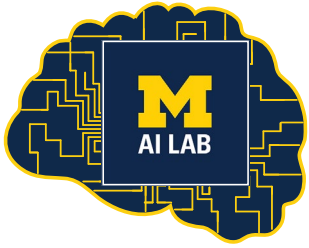

# University of Michigan
# AI LABORATORY

**WHITE PAPER**

# Evaluation Framework for AI Systems in "the Wild"

**LIFECYCLE EVALUATION**

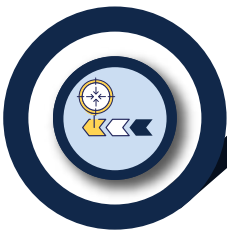

**HOLISTIC EVALUATION**

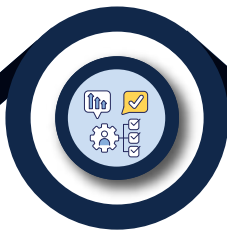

**HUMAN AND AUTOMATED EVALUATION**

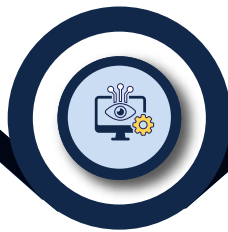

**DYNAMIC AND ADAPTIVE EVALUATION**

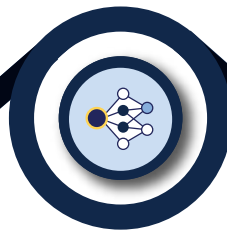

**SAFETY AND SUSTAINABILITY EVALUATION**

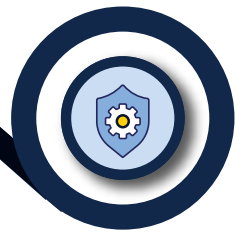

# Authors:

Sarah Jabbour[1]
Trenton Chang[1]
Anindya Das Antar[1]
Joseph Peper[2]
Insu Jang[2]
Jiachen Liu[2]
Jae-Won Chung[2]
Shiqi He[2]

Michael Wellman[3]
Bryan Goodman[3]
Elizabeth Bondi-Kelly[3]
Kevin Samy[3]
Rada Mihalcea[3]
Mosharaf Chowdhury[3]
David Jurgens[4]
Lu Wang[4]

[1] *Lead student contributors*
[2] *Student contributors*
[3] *Senior contributors*
[4] *Corresponding authors: {jurgens, wangluxy}@umich.edu*

# Table of Contents

# Key Takeaways

- Models will continuously change as technology evolves and we need outcome-oriented evaluations, rather than model-oriented, that reflect the impact of AI on users and society.

- Human judgment is essential to AI evaluation but automated methods are needed for scaling evaluation to the range of capabilities.

- AI Evaluation is multifaceted and much of the community has only focused on single, static facets like benchmarks. Holistic evaluation is needed that reflects the impact on people, society, and the environment.

- The evaluation process needs to be clearly communicated to all relevant stakeholders so that its findings are understood and accepted.

- Evaluation is not just a machine learning problem: evaluation requires perspectives from different stakeholders, practitioners, and domain experts, which requires interdisciplinary cooperation.

- Metrics need to be identified in the context of application deployment and practitioners need to understand the tradeoffs among the factors that affect the performance.

# Executive Summary

Generative AI (GenAI) models have rapidly advanced and they have become integral across a wide range of industries, including healthcare, finance, education, and entertainment. However, the processes and methodologies for evaluating these models have not kept pace with their widespread deployment. Traditional evaluation methods heavily rely on standardized benchmarks and fixed datasets. Though offering a controlled and reproducible way to assess specific model capabilities, these static evaluations often fail to capture how models perform in real-world scenarios, thus leading to a gap between lab-tested performance and real-world outcomes. As a result, the current state of evaluation for GenAI models remains limited. This highlights the need for more holistic, dynamic, and continuous evaluation approaches.

This white paper offers a comprehensive framework for **in-the-wild GenAI model evaluation**, where they are deployed in real-world settings with diverse and evolving inputs.
**Practitioners** will find guidance on designing evaluation methods that reflect the models' capabilities and behavior over time. **Policymakers** are provided with recommendations to draft AI policies that focus on the outcomes and societal impacts rather than specific model design choices or characteristics.
**Researchers** can benefit from understanding how their models will be evaluated in practice, particularly in dynamic and high-stakes environments.
**Sponsors from funding agencies** will be better equipped to allocate resources to support dynamic, robust, and scalable evaluation methods that match the complexity of real-world applications where the models are deployed.

Second, GenAI models are increasingly used in high-stakes applications, where errors or biases can lead to significant real-world consequences, such as in healthcare or financial decision-making. This white paper identifies the limitations of current evaluation practices and emphasizes the need for a shift toward adaptive and outcome-oriented evaluations that consider not just accuracy or efficiency, but the broader impact of AI on users and society.

To address these challenges, this white paper outlines several key recommendations. First, it advocates for holistic evaluation frameworks that integrate multiple metrics, including performance, fairness, and ethical considerations. Second, it underscores the importance of incorporating human judgment into the evaluation process, while also recognizing the need for automated methods to scale evaluations across a range of model capabilities. Furthermore, the paper stresses the importance of transparency in the evaluation process, to allow the stakeholders to understand and trust the outcomes of AI evaluations. Finally, it highlights the need for interdisciplinary and adaptive evaluation methods that draw on insights from various fields and adjust to the specific contexts in which models are deployed. Two case studies are provided to illustrate how these recommendations can be practically implemented.

In conclusion, this white paper provides actionable recommendations for improving the evaluation of GenAI models in real-world scenarios. Key takeaways include the need for continuous, dynamic, and outcome-oriented evaluation methods; the integration of human judgment with automated evaluation processes to balance scalability and context-awareness; and the development of transparent and interdisciplinary approaches that foster trust among all stakeholders. By adopting these recommendations, the AI community can progress in a direction where GenAI models are not only technically proficient but also ethically sound, reliable, and impactful in diverse application environments.

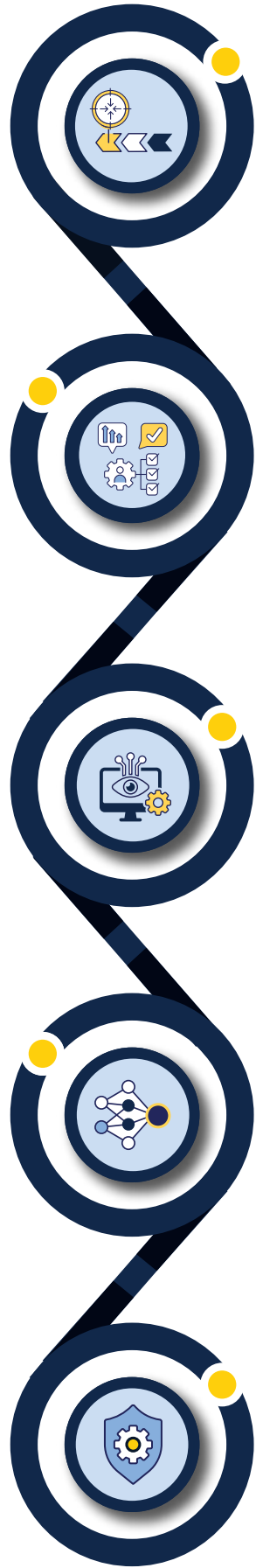

## LIFECYCLE EVALUATION

Continuously monitor AI from development to deployment

Evaluate outcomes that reflect societal impacts

Regularly re-evaluate deployed systems on new benchmarks

## HOLISTIC EVALUATION

Move beyond benchmarks to real-world performance

Assess AI with user interactions, societal impact, and ethical implications together

Collaborate across disciplines to comprehensive assess

## HUMAN AND AUTOMATED EVALUATION

Balance nuanced human judgment with scalable automated evaluations

Leverage diverse stakeholder expertise

Incorporate qualitative insights with quantitative data

## DYNAMIC AND ADAPTIVE EVALUATION

Recognize that benchmarks grow stale and may overestimate performance

Design evaluations to reflect real-world complexities and evolving needs

Continuously adapt and update benchmarks to prevent stagnation

## SAFETY AND SUSTAINABILITY EVALUATION

Proactively measure safety and risk

Include energy efficiency and environmental considerations as key metrics

Evaluate potential biases and unintended impacts for individual and society

# Introduction to GenAI Evaluation

GenAI techniques have become a cornerstone of modern AI development and have been powering innovations across industries such as healthcare, finance, education, entertainment, and beyond. From generating human-like text in conversational agents to creating artwork and code, GenAI models have significantly expanded the realm of what traditional AI can achieve. The versatility of GenAI models has enabled breakthroughs in areas that were once thought to be the sole domain of human expertise.

For instance, GenAI models have been used to assist in drug discovery and accelerate the development of new treatments. In finance, they help automate tasks such as fraud detection and risk assessment. In education, GenAI powers intelligent tutoring systems that provide personalized learning experiences by adapting to each student's needs and pace. In the entertainment industry, these models have been transforming the creative process by generating new music tracks, creating realistic visual effects for films, and assisting in the production of video games. These examples highlight how GenAI models are pushing the boundaries of what AI can accomplish.

As GenAI models become deeply embedded in real-world applications, properly evaluating them becomes a critical step to understanding their performance, safety, and impact. Currently, most evaluations focus on well-defined metrics, such as accuracy or F1 score, and rely on standardized benchmarks[1]. These benchmarks are valuable for measuring specific capabilities of models, but they offer limited insight into how models perform in the unpredictable and diverse conditions of real-world uses. Moreover, because these evaluations emphasize static performance and isolated tasks, they can become ineffective and obsolete as models evolve—often failing to predict real-world performance in diverse deployment contexts.

---

[1] A **benchmark** is a type of test where models perform a task like answering a multiple choice question for different examples. Each benchmark is paired with a metric for measuring how well a model does on the task relative to its answers. More complicated tasks may involve summarizing passages or answering questions, which require more complicated metrics.

> *What is missing from the current GenAI evaluation practices is a holistic, continuous, and dynamic approach to testing model performance in real-world scenarios.*

As GenAI models are deployed in critical fields like healthcare and finance, evaluating their long-term reliability, fairness, and transparency becomes as important as assessing their immediate accuracy. Moreover, evaluating these models in isolation from their deployment context is insufficient. A more comprehensive evaluation framework must consider how these models interact with users, evolve over time, and impact broader societal outcomes. Without these considerations, there is a risk that AI systems will fail to meet real-world demands or even exacerbate existing inequalities and risks in sensitive applications.

The purpose of this white paper is to address these gaps by providing a framework for evaluating GenAI models in real-world scenarios. Practitioners will find guidance on designing holistic, dynamic, and continuous evaluation plans that account for how models evolve and adapt over time. Policymakers can use the document to develop regulations that remain robust in the face of rapidly advancing AI technologies, focusing on outcomes and societal impacts. Researchers will gain insights into how their models might be evaluated beyond traditional benchmarks, to better align their work with real-world demands. Finally, sponsors from funding agencies will be equipped to make informed decisions about scaling resource investment in line with the evaluation needs of advanced AI systems.
Through this effort, we aim to ensure that AI systems not only excel in lab settings but also perform reliably, ethically, and effectively in the diverse environments they will encounter in the real world.

# Existing Evaluation Methods and Limitations

Using standard benchmarks and automatic metrics that measure model performance against references has been the standard practice for evaluating AI models across various domains such as natural language processing (NLP) and computer vision (CV). A standard benchmark often consists of static datasets to evaluate the performance of models on specific capabilities. There are several reasons why this framework has come to dominate model evaluation. First, with standard benchmarks, consistent, objective, and reproducible evaluation across different models becomes possible. Second, benchmarks are often constructed to highlight the major challenges for a domain or an application, so that they can guide the community to focus on the most critical problems. Moreover, evaluation on standard benchmarks also enables tracking of research progress on specific problems and the identification of state-of-the-art techniques and models.

On the other hand, benchmarks also have several limitations. First, models may overfit to perform well on benchmark tests rather than generalizing to real-world tasks. Therefore, the improvements made in a limited number of benchmarks can be narrow and do not translate into broader applicability. Moreover, high-quality benchmarks require careful design to reflect the complexities and a wide range of scenarios in real-world problems, while existing datasets often do not cover the full spectrum of capabilities needed in practical applications. Furthermore, with the rapid iteration of GenAI models, performance on static benchmarks quickly reaches saturation. Therefore, it is crucial to continually update and expand these benchmarks to better evaluate the evolving capabilities of these models.

## Takeaways

- Current benchmark-based evaluation methods provide consistent, objective, and reproducible assessments of AI models, but they must be carefully designed to capture real-world complexities and require continuous updates to stay relevant amidst rapidly evolving model capabilities. This ensures that evaluations remain comprehensive and applicable to practical applications.

# What is Being Evaluated?

No single evaluation can measure the full capabilities of AI systems. As a result, practitioners have proposed multiple complementary tests for different qualities of a system's output and the impact that system might have.

## In-the-lab evaluation

*In-the-lab evaluation* is a widely used approach to assess GenAI models, through the use of standardized benchmarks and datasets. This method allows practitioners to systematically measure a broad range of capabilities. In the aggregate, these varied datasets and benchmarks provide a holistic view of a model's general-purpose strengths and weaknesses when tested on a wide array of difficult tasks ranging from coding (HumanEval), math reasoning (GSM8K), and situated video reasoning (STAR), as well as when pushing the boundaries of model capabilities, such as handling long inputs (LongBench) or handling multiple modalities (MMMU).

By offering controlled conditions and standardized metrics, in-the-lab evaluations enable reproducible experiments and fair comparisons across different models. This structured approach allows for a clear understanding of how well models handle specific types of problems. However, the controlled nature of these benchmarks means they may not fully capture how models perform in dynamic, real-world environments, where tasks can be more complex and less predictable. To address this gap, complementary evaluation methods—such as user-driven evaluations or real-world testing—are often necessary to provide a fuller picture of a model's practical utility and adaptability.

### Takeaways

- Standardized benchmarks enable practitioners to evaluate key model capabilities in a reproducible manner that allows for comparison with other models.

- While extremely valuable, in-the-lab evaluation alone cannot fully represent model performance in dynamic, real-world settings.

## Human capability-specific evaluation

Tests such as the GRE, MCAT, or LSAT are designed to evaluate human knowledge and skill. These tests often evaluate significant domain knowledge, advanced reasoning capabilities, and the ability to generalize one's knowledge to new and dynamic situations.
AI practitioners have adapted human qualification exams to also test AI capabilities. Advanced LLMs such as GPT-4 and other state-of-the-art models have scored far above the human average on competitive professional exams, such as the CFA and medical licensing examinations, that require answering multiple-choice questions as well as generating essays and short responses.

However, when evaluating humans, these exams typically involve other humans judging the answers, such as attorneys judging responses to the bar exam. As a result, these essay-based answers introduce new complexities for how to measure AI response quality in a replicable setting. More broadly, these exams are taken within a particular context, e.g., board exams at the end of medical school, which often implies a certain level of training or expertise by the person. Even if an LLM performs well on a professional exam, this doesn't necessarily translate to the same real-world expertise that a human has.

## Takeaways

- Exams for human expertise can be adapted to assess AI models but evaluating the AI models' answers the same as humans may not be feasible.

- Correctly answering these exams does not generally imply models have the same abilities as humans who successfully take these exams.

# In-the-wild evaluation

*"In-the-wild" evaluation* focuses on the performance of a model for a specific, practical use case. Such evaluations are tailored to the priorities of stakeholders relevant to the use case. In-the-wild evaluation extends beyond model performance on fixed benchmarks/datasets and considers how a model's outputs interact with existing real-world workflows.

While in-the-lab evaluation via curated, widely-available benchmarks is important for approximating the general capabilities of GenAI models in a reproducible and comparable manner, there are differences between benchmark performance and real-world model usage. First, there is a potential validity mismatch between benchmarks and practice: metrics such as question-answering accuracy or test scores may not directly correspond to the priorities of those who interact with the model. Furthermore, benchmarks may not have full coverage over specific, real-world failure modes that model users might experience in practice, which in-the-wild evaluation can help surface prior to wider adoption.
Lastly, a GenAI may serve a variety of roles in practice: e.g., Gen AI output might be included only as one piece of supporting evidence among many, or be used to generate a "draft decision" that is later confirmed/revised by a human. The use of GenAI in these complex workflows makes its individual evaluation complicated. Many of these limitations are shared by older AI models, and GenAI models are no exception.

While there are many criteria for designing an effective in-the-wild evaluation protocol, we suggest a few overarching themes:

o **Defining contextually appropriate performance metric(s).** A good metric should be defined based on the intended task and context. For example, in a healthcare diagnosis task, beyond diagnostic accuracy, one might consider downstream patient outcomes (e.g., risk of readmission) or cost savings from using a GenAI model. One may also consider evaluating multiple metrics in parallel to assess trade-offs between priorities (e.g., privacy vs. fairness vs. top-line performance).

o **Capturing unintended impacts.** Evaluations should be conducted using data similar to that a GenAI model would encounter in practice, in contrast to out-of-the-box benchmark data, which may or may not be representative of a practical setting. While "similarity" is often challenging to assess in practice, practitioners should exercise due diligence in curating data for evaluation, e.g., accounting for variation across population subgroups (e.g., racial, gender, and other biases in model outputs).

o  **Considering workflow effects.** In practice, decision-making processes are often complex. Thus, GenAI model outputs could have various levels of impact on a decision-making process (e.g., is the output one piece of evidence among many, or a "draft decision?"). A successful in-the-wild evaluation should holistically consider the effect of a model output on the decisions rendered by a specific process or workflow.

These desiderata are not independent. For example, considering workflow effects could motivate metrics based on comparing outcomes of a decision-making process with vs. without a GenAI model. In practice, these criteria can be implemented through user studies that evaluate how practitioners use GenAI tools or shadowing studies, where GenAI outputs are made visible for evaluation without being directly acted upon.

## Takeaways

- In-the-wild evaluation should be tailored to bridge the gap between in-the-lab benchmarks and a practical, scoped use-case of a GenAI system.

- Important desiderata for in-the-wild evaluation include ensuring that the relevant metrics align with real-world priorities or practices and capture potential unintended impacts of GenAI system usage.

# Who Evaluates and How?

Achieving comprehensive evaluation of AI models involves balancing standardized automated methods with human insights and leveraging the expertise of diverse stakeholders to assess both technical performance and real-world impact.
While benchmark-based evaluations provide measurable performance metrics, human-centered evaluation methods are crucial for understanding how models align with user expectations, values, and societal norms. These considerations are particularly important for generative AI models, such as large language models (LLMs), which are increasingly being explored as evaluators in different tasks.

## Benchmark-based Evaluation

Benchmark-based evaluation of AI models involves investigating machine learning (ML) models using standardized datasets and tasks to assess AI capabilities and limitations, compare different models, and ensure they meet the required standards for deployment. Benchmarks provide a structured environment to evaluate model predictions against ground truth labels using predefined quantitative performance metrics, such as accuracy, precision, recall, and other specialized measures depending on the task. However, the utility of this approach is highly dependent on the quality of the evaluation data and does not inherently provide explanations to help model engineers understand the reasoning behind the model's decisions. Although there are some open-source ML benchmarking tools available, they are often tailored to assess specific types of lightweight ML models with limited resources.

Benchmarks are typically designed and constructed by a diverse group of contributors, including but not limited to model engineers, domain experts, academic researchers, industry experts, and standardized organizations. However, their objectives of benchmark design extend beyond simple model evaluation. They also aim to establish a standardized basis for model comparison and identify gaps in existing evaluation methodologies.  By setting clear evaluation criteria, benchmarks help them assess progress toward solving specific challenges and uncover areas where models may struggle, such as dealing with edge cases or adapting to new environments. Also, they select and curate benchmarks with specific objectives in mind, such as evaluating model accuracy, bias, alignment, fairness, and generalization ability, and often target distinct AI tasks, such as object detection, machine translation, or question answering. After identifying the target task, they collect and annotate data from relevant sources, such as publicly available datasets, domain-specific repositories, or collecting annotations from a large number of laypersons, which is known as crowdsourcing.
In some cases, they also generate synthetic data to create controlled environments for evaluating model capabilities and limitations. Therefore, they must understand and carefully consider the choice of benchmarks when selecting models for their specific application domains and making informed decisions.

# Human-centered ML Evaluation

Evaluating machine learning models and AI-based systems requires a comprehensive understanding of their performance, capabilities, and limitations in real-world contexts. A human-centered approach to evaluation, especially for large language models (LLMs), integrates quantitative metrics with qualitative insights and human input to ensure these models align with user needs and expectations. While this approach often prioritizes factors such as accuracy, usability, and fairness of the model, such evaluations tend to be slow and subjective, relying on human intervention and influenced by individual biases, which can affect the overall reliability and consistency of the assessment.

## Tradeoffs between automated and human-centered evaluations

*Evaluating machine learning (ML) models requires carefully balancing automated testing with human-centered evaluations, where humans actively intervene and probe the models.*

However, each approach offers distinct advantages and limitations. Automated tests, which encompass performance metrics, stress tests, and error analysis, provide efficiency, scalability, and consistency, making them well-suited for large-scale, repeatable assessments. However, they often cannot capture the complexity of real-world interactions or how the model performs in settings different from those in which it was developed, potentially overlooking critical contextual factors. In contrast, human-centered evaluations involve direct human engagement to assess aspects like usability, fairness, and potential biases. While these evaluations provide valuable, context-aware insights, they are often time-consuming and costly. Also, given its subjectivity, human-centered evaluation is limited to the specific characteristics and biases of the group performing the evaluation and can be difficult to scale up to encompass a comprehensive range of experiences. Thus, a comprehensive evaluation strategy must integrate the rigor of automated testing with the nuanced ethical considerations and real-world adaptability offered by human intervention.

# Roles and expertise of stakeholders in human-centered evaluation

Different stakeholders play critical roles in evaluating the model throughout various stages of the ML model lifecycle, from the early stage of model development to deployment and beyond. Understanding their roles and expertise is essential for conducting comprehensive evaluations. **Model engineers** are the first in line to develop and evaluate ML models. While some are highly trained computer scientists with formal expertise in model development and evaluation, many come from diverse coding backgrounds drawn by the growing demand for ML and generative AI (genAI) technologies. This democratization fosters innovation but raises concerns, particularly in high-stakes settings, about the lack of deep expertise in underlying model mechanics and potential downstream consequences. In many cases, developers may focus on surface-level functionality without fully considering ethical implications, bias, or long-term risks. Thus, robust evaluation is essential to ensure that ML models are not only functional but also robust and ethically sound, especially when deployed in high-stakes domains.

There are also end-users (i.e., domain experts, policymakers, and consumers) who may not have computer science-savvy expertise but seek to understand and explore model decisions. **Domain experts** are professionals with knowledge of specific domains where the models will be deployed. They may or may not have experience interacting with ML-based assistive decision support systems as part of their workflow. **Policymakers** are individuals or organizations responsible for setting rules and regulatory standards for these models' deployment in decision-making. While their role is to ensure compliance with legal and ethical standards, they may not always have a deep understanding of how these models are developed or function. Recent legislative debates illustrate that some policymakers may propose regulations that reflect existential concerns rather than the technical realities of AI systems. Also, there are **consumers** who directly interact with or are affected by ML-based decision-making systems. They also have relevant domain knowledge due to their interactions with the systems but not necessarily related to their professional and business activities.

However, most existing ML evaluation systems require computer science-savvy expertise (e.g., ML, task, and UI expertise) and primarily target model engineers. Thus, such tools do not directly support other end-users who may not possess or seek to have such expertise in evaluating the models they interact with.

# Takeaways

- Benchmark-based ML evaluation relies heavily on standardized datasets and tasks to assess model performance using predefined quantitative metrics.

- Human-centered evaluations involve users with various expertise (e.g., ML expertise, domain knowledge, policy expertise) in the evaluation process, providing context-aware insights, but they are slower, more subjective, and influenced by evaluator biases.

- While these end-users (e.g., domain experts, policymakers, and consumers) offer essential domain knowledge and regulatory insights to ensure ML models are effectively integrated into workflows, they may not possess, nor do they seek computer science-savvy expertise that most existing evaluation systems primarily focus on.

- We need evaluation systems that support not only model engineers but also end-users without computer science expertise to evaluate ML models.

# LLM as a Judge

In some settings such as essay writing, code review, or other writing-like tasks, there are many different correct answers to the same question due to stylistic differences.
Scoring these outputs with traditional metrics is difficult due to the variety of responses. Ideally, human evaluators are the gold standard for assessing quality given the varied nature of the model output. However, such evaluations can be expensive to solicit and hard to replicate. As a result, LLMs have been used in place of humans to evaluate the quality of model outputs. Because of their powerful language understanding abilities, treating LLMs as evaluators allows practitioners to design nuanced evaluation metrics for evaluating far more complex LLM outputs than possible with previous metrics like Accuracy.
These LLM-as-judge systems can be used to evaluate many types of outputs and behaviors, such as conversational fluency.

When effective, LLMs can serve as fast and replicable evaluators for a variety of settings, particularly when prompted with detailed and objective rubrics for scoring.
Nevertheless, using LLMs to evaluate quality can introduce multiple biases. Models are known to prefer certain styles of output due to their training, potentially biasing evaluation when the same model is used to generate output and evaluate. Further, models are still limited by the complexity of the task relative to the model's language understanding abilities, particularly for tasks that involve very long passages of texts or those that involve highly detailed scoring rubrics. Good practices for using an LLM to evaluate the output of other LLMs include verifying that an LLM's scores align with human judgments, using one or more different LLMs to evaluate that are different from the model used to generate so that the model doesn't prefer its own outputs.

## Takeaways

- LLMs can provide fast evaluations across various tasks, particularly for generation tasks when detailed rubrics are used to guide the evaluation process.

- The use of LLMs can introduce biases and preferences for certain language styles, and challenges with handling different input lengths, modalities, or complex tasks.

# Other Considerations

## Objective vs Subjective Evaluation: Alignment with Human Perspectives, Values, and Ethical Standards

Some subjective questions may have different answers that vary based on the person answering. For example, what is considered offensive may vary slightly based on age, gender, values, and lived experience of a person. The answers to other types of questions depend on the values of the person, e.g., the correct action to take in a particular scenario. In these settings, traditional evaluation metrics are ill-suited because they do not capture the meaningful differences in what is considered "correct" relative to who is judging the answer.

AI models can be evaluated with respect to how much their answers **align** with the answers of specific groups or values, and with respect to how well they are capable of taking a specific perspective (e.g., from a specific group) when instructed to in a prompt. **Practitioners and developers** should be aware of how well the views and preferences in an AI model's output align with the communities that will use them—especially when these communities' views may not have been incorporated during model training (for example, via [participatory design](#)). While AI decision alignment with user preferences is important, alignment goals may also extend beyond individual preferences to broader societal norms and ethical standards established by subject matter experts in the domain.
Effective human-AI alignment evaluation will also require substantial data on subpopulation preferences to understand how well and when AI outputs match these subpopulations' preferences.

### Takeaways

- Models need to be evaluated according to how they can simulate different human preferences when asked to assume the perspectives of different subpopulations.
- Models should also be evaluated on how effectively their behavior can be shaped to align their decisions with user expectations, values, and established ethical standards.
- Moreover, evaluating the effectiveness of human-AI alignment will also require substantial data on subpopulation preferences.

# Safety-focused Approaches

Currently, AI models are often developed without a thorough understanding of their safety implications, even though risks typically materialize only after deployment. A proactive approach requires developers to demonstrate the safety of their AI models when starting development. This shift not only mitigates potential hazards but also encourages research focused on predicting and controlling AI safety throughout the development process.

According to many lab insiders—as well as [best practices](#) from organizations that develop high-impact technologies—this process is counterproductive. Building powerful systems with unknown safety properties, and then determining their safety properties only after development or use, exposes critical infrastructure and the public to severe risks. Instead, developers of the most powerful AI systems should build a clear and compelling safety case—demonstrating with high confidence the safety and security properties of their models—at the design stage before development occurs. This common-sense measure aligns with what Americans expect from high-stakes research. High-risk activities should not proceed without proving they can be executed safely.

To effectively implement a safety-forward policy, three key elements are necessary. First, there needs to be a clear regulatory framework for all companies involved in the AI supply chain. This includes establishing a licensing system that outlines the requirements for safe AI development. Second, a dynamic regulatory body is essential to keep pace with the rapidly evolving AI landscape. This agency should have the flexibility to update regulatory standards based on advancements in AI technology, ensuring that safety measures remain current and effective. Third, a comprehensive liability framework is needed to hold companies accountable for non-compliance. This framework should impose consequences for breaches in safety protocols and any harm caused by AI systems, thereby incentivizing adherence to safety standards.

Adopting a safety-forward approach to AI development accelerates the integration of safety and security practices, aligning regulatory oversight with the varying levels of risk associated with different AI models. This strategy fosters an environment where innovation and safety coexist, ensuring that powerful AI technologies are developed responsibly. By prioritizing these safety considerations, we can leverage the benefits of AI while safeguarding society against its potential risks.

<div align="center">**Takeaways**</div>

- Current AI models are often developed without adequately understanding their safety implications, leading to risks manifesting post-deployment.

- Developing powerful AI systems with unknown safety properties and assessing them only after deployment exposes critical infrastructure and the public to severe risks.

# Practical Systems Evaluation

AI models require power, hardware, and cooling to operate. These physical requirements may be substantially different across models and are an important practical dimension for evaluation.

## Holistically evaluating systems for ML training

*Evaluating GenAI models effectively requires not just measuring end results but also understanding the efficiency, scalability, and sustainability of their entire lifecycle: from training to deployment.*

Evaluating GenAI models in practice requires a broader approach than traditional machine learning assessments, which primarily focus on accuracy. While accuracy is important, practical evaluation needs to capture the full lifecycle of model development, particularly the significant resources required for training and ongoing updates.
A common misconception is that once trained, a model can be used indefinitely with minimal additional costs. However, most production GenAI models are continuously retrained to incorporate new information, address knowledge limitations, and improve performance, resulting in recurring computational and infrastructure demands.

## Evaluating deployed GenAI systems

Once trained, GenAI systems are typically deployed for use by people and applications. This phase, known as the inference phase, requires continuous resources to host the models and process requests. Evaluating this phase of the model lifecycle focuses on the experience of the end-user and the resources required for service. Traditional metrics for evaluating models include throughput as how many requests can be processed per second, time-to-first-token for how long a model takes from receiving a request to generate the initial part of its response, and total latency for how long requests take.

While training GenAI systems requires substantial resources, deploying these systems for inference only typically requires far less. For some GenAI models, techniques like quantization can drastically reduce deployment requirements, enabling practitioners to run the models on smaller systems or run more models on larger systems. Models can be evaluated by the scale of resources needed for deployment, which directly affects consumers' experiences and the costs required to operate the GenAI system.

## Power, energy, and sustainability considerations

Today, the IT industry is estimated to account for about 8% of global electricity demand. Within the IT industry, AI workloads, particularly those involving LLMs, consume a significant portion of resources due to massive data scales, large model sizes, intensive use of power-hungry GPUs, and rapidly growing real-world adoption. As a concrete example, Amazon trained a 200 billion parameter LLM which reportedly used around 11.9 GWh of energy—enough to power more than a thousand average US households for a year.

This level of energy use has several implications. It increases energy bills, raises sustainability concerns, and adds to the carbon offsetting costs for companies aiming for Net Zero emissions. Additionally, the high demand for power has become a key bottleneck in data center construction and operation. This not only increases the costs and lead times for building new data centers but also limits where they can be feasibly built, especially in areas with unstable or insufficient power supplies. *Components in the AI lifecycle must be evaluated in terms of how much energy they use, and how efficient they are in their use of energy.*

The evaluation and enhancement of energy efficiency, like any other optimization metric, requires precise measurements as a prerequisite. Yet, existing literature often falls short by providing coarse estimations of energy consumption or carbon emission, e.g., using the hardware's maximum possible power consumption as an estimate for actual power consumption. As the discussion around AI energy consumption moves beyond mere awareness, both the open-source/research community and policymakers must advocate for precise measurements of realistic workloads to better optimize for energy. To this end, there are already existing tools for energy measurement and optimization like [Zeus](#).

Energy consumption is just one metric among many important systems metrics, including speed and computing resource requirement, and the trade-off space created by all such metrics is complex. For instance, a smaller AI model may not always be more energy-efficient than a larger one, depending on the model's architecture and the type of service it is used in. For instance, in multi-turn conversations, a smaller LLM might use less energy to generate a single word but require many more turns to reach the desired result—eventually consuming more energy than a larger LLM. Therefore, it is important to identify the end metric of the specific AI application under evaluation and to explore and understand the trade-off space that involves energy consumption as one dimension. As service metrics and the trade-off space are highly dynamic, policies should be general or flexible enough to quickly accommodate changes in the landscape.

## Takeaways

- Today, energy is a scarce computational resource that requires careful optimization.

- Open-source/research communities and policymakers should push for precise measurement with the goal of optimizing.

- It is important to understand and build around the complex and dynamic trade-off space including energy and other systems metrics.

- AI models are not static assets; they require ongoing investments in infrastructure and refinement. Policymakers should consider regulatory frameworks that balance innovation with economic and environmental sustainability.

## Data processing/selection vs. model performance

LLM training datasets are often large in size (e.g., FineWeb contains 15T tokens), containing a diverse and comprehensive collection of data points to ensure the model can learn robustly. In contrast, LLM evaluation datasets are much smaller (e.g., MMLU contains only 15K questions) but tend to be more representative of the problem domain. They should cover various aspects of the data to provide a reliable measure of model performance and are often curated to evaluate specific cases, tasks, and domains.

Despite the smaller data size of evaluation datasets, evaluating LLMs on a number of evaluation benchmarks with thousands of samples could still be environmentally and financially costly, especially because modern LLMs consist of billions of parameters. For example, evaluating the performance of a single LLM on the Holistic Evaluation of Language Models (HELM) benchmark could cost thousands of GPU hours. Furthermore, evaluation might also be performed many times when developers monitor checkpoints during pre-training and explore different hyperparameters (e.g., learning rates) or training settings (e.g., prompting).

Data selection/sampling is an important strategy for efficiently speeding up LLM evaluation by reducing the dataset size while maintaining its representativeness. One common approach is random sampling, which means selecting a subset of data points at random. Stratified random sampling goes a step further by dividing the dataset into distinct groups based on certain scenarios and then randomly sampling from each group proportionally. This helps in preserving the distribution of different categories within the data. Additionally, cluster sampling groups data into clusters based on their similarities. The data samples inside one cluster are expected to have the same prediction result (i.e., correct or incorrect). While recent studies propose more advanced sampling strategies, performing data selection on evaluation datasets can still result in an estimation error. Developers should carefully consider the sampling strategies together with an appropriate sampling ratio.

## Takeaways

- Evaluating modern LLMs can be costly in terms of resources and time. Despite being smaller in size, evaluation datasets require significant computational resources due to the complexity and size of the models.

- Even with advanced sampling methods, there can still be challenges in accurately estimating model performance due to potential estimation errors.

# How Long is the Evaluation Relevant?

Creating and validating evaluation benchmarks involves significant effort, and this overhead is a considerable factor in real-world settings. Without updates, benchmarks can become outdated, failing to accurately reflect the challenges that models face in real-world applications. Automation also plays a crucial role in modern benchmarking processes, allowing for scalable, efficient, and adaptable evaluation to keep pace with the rapid growth of GenAI capabilities. In addition, GenAI-assisted automation factors into modern benchmark development, and it is important to understand the tradeoffs of these methods.

*As GenAI continues to grow in capability, it becomes necessary to ensure that evaluation benchmarks remain relevant and effective.*

## Dynamic Evaluation + Benchmark Automation

As Gen AI capabilities rapidly grow, it is crucial to regularly update and iterate on evaluation benchmarks to ensure they remain challenging and relevant to the field. Older benchmarks can be uninformative or even misleading, as they fail to reflect the new and growing challenges experienced today. Similarly, outdated benchmarks intended for weaker models can suffer from performance saturation, leading to poor performance differentiation. As such, there is a need to adopt a dynamic and rolling evaluation methodology that continuously checks and adjusts benchmarks to reflect emerging needs and more performant models. A component of this is understanding and reacting to signals that indicate benchmark limitations, including performance saturation, temporal incongruence, and data contamination. These measures can allow us to work towards robust and meaningful evaluation, accurately reflecting the capabilities of modern LLMs.

Automated tooling can ease the benchmark development process, reducing cost overheads and cognitive burden for benchmark maintainers. Recently, AI-in-the-loop methods have proven very beneficial in assisting in benchmark development. However, these approaches are not without risks; there are notable tradeoffs between human-generated and synthetic benchmarks, including the time required to create examples, the quality of the benchmarks, and their domain coverage and diversity. In addition, the intended evaluation setting also influences these decisions; many real-world industry applications can require incorporating proprietary or niche knowledge requiring specialized expertise during annotation.

# Evaluating the Evaluations

To evaluate whether a benchmark is still relevant to quantifying models' performance, the first thing is to check the quality of the benchmark, e.g., whether there are any errors and the diversity and complexity of the samples. Second, if models consistently achieve high scores or there is little variance in performance among different models, the benchmark may no longer effectively differentiate between model capabilities. Third, it is also important to measure whether the tasks or samples in the benchmark still reflect the challenges in the current real-world applications. Fourth, the models' performance on the dataset should correlate with human judgment of the models' usefulness in real-world applications.

To evaluate whether an evaluation method or a metric can correctly quantify models' performance, it would be important to investigate the reliability and standards used in these evaluations. Take language model evaluation as an example, several factors may affect evaluation results, including the prompts (such as tone, detailedness, and formats) and the number of shots and examples used when in-context learning is applied.
Moreover, it is critical to know whether the metrics evaluate the essential properties of the tasks and applications. Finally, when another model is used as the evaluator, it is important to ensure the same model is used to produce the assessment and the model should be made publicly available.

As datasets age, the risk of data leakage becomes increasingly significant. Data leakage occurs when information from the test set is included in the training process, leading to inflated performance metrics that do not accurately reflect a model's true capabilities. This risk grows when the dataset is available for a long time, as researchers and developers may inadvertently or purposefully expose models to test data during training or fine-tuning. This reduces the dataset's ability to serve as a robust measure of generalization, as models may exploit dataset-specific quirks rather than developing genuine capabilities applicable to broader, real-world scenarios. Therefore, continuous refreshment and diversification of datasets are crucial to maintaining their integrity and ensuring that they provide a true measure of model performance.

# Takeaways

- Evaluation benchmarks and datasets must be **regularly updated** to prevent stagnation, minimize data leakage, and ensure they remain relevant to evolving tasks and model capabilities.

- Benchmark creation is **expensive** and it is important to consider how **LLM-assisted automation** can benefit this process.

- It is essential for benchmarks to maintain a **diverse and complex set of samples** to accurately reflect the range of challenges presented in real-world applications.

- Benchmarks should measure aspects of model performance that **correlate well with human judgment** of utility in practical scenarios, ensuring that improvements on benchmarks translate into real-world benefits.

- Evaluation metrics and methods must be **reliable, reproducible, and adhere to consistent standards.**

- When using another model as an evaluator, it is important that the **evaluator model is made publicly available** to ensure transparency and reproducibility in evaluations.

# Case Studies:

Real-world examples of AI in practice highlight the necessity and complexity of evaluating generative AI. Here, we highlight two case studies on how one might consider implementing the themes discussed above.

## Health AI: Clinical Note Summarization

AI has shown promise in enhancing patient care, e.g., through clinical tasks like interpreting medical images or automating scribing. The task of a clinical note summarization model is to generate a concise after-visit summary to send to a patient, given a transcript of a physician-patient conversation. This summary may feature action items, such as information about follow-up appointments or medication instructions. Accuracy and readability are critical to improve patient adherence to treatment regimens and reduce administrative burdens.

## What is being evaluated?

Clinical summarization tools can be evaluated against static benchmarks that use human-generated summaries as a reference. Such an automated evaluation is especially useful when choosing from a set of potential models to use. However, automated summarization metrics come with limitations (e.g., evaluating performance with an LLM or using word overlap-based metrics) and may not align with clinical utility or fully demonstrate the capabilities of the model.

Practical evaluations need to look beyond static benchmarks. For example, consider a procurement officer at a hospital deciding between models to recommend to the hospital administration. The procurement officer may prioritize metrics such as time saved by clinicians, improvement in patient outcomes, and cost savings to hospital operations—none of which are captured by summary quality evaluations. Since these metrics are defined in the context of a hospital, they could be difficult to anticipate during the development of a GenAI model. As such, models that perform the best on benchmark metrics, may not necessarily translate to performing the best in terms of metrics important to the hospital.

In summary, the best model for a GenAI researcher may not be the best model for a hospital. We encourage researchers, regulators, and domain experts interfacing with AI models to consider discrepancies between notions of performance and to collaborate on bridging the gap between benchmark performance and real-world adoption.

## Who should evaluate?

In the context of evaluating a GenAI model in healthcare, many different perspectives could offer unique insights into the clinical utility of the model. For instance, ML researchers and engineers have the expertise to build systems and evaluation tools, especially in terms of building and scaling automated evaluations. Thus, they can work closely with clinical experts to understand their needs and implement such evaluations accordingly. Hospital administrators and clinicians could also contribute their understandings of how a potential GenAI model would fit into hospital operations, helping define evaluation targets and outcomes. For models with patient-facing implications, e.g., the generation of after-visit summaries, understanding how model outputs interface with a wide variety of patients' experiences with the healthcare system may be important to consider as well. Equally important are policymakers and legal experts, who could help contextualize how this system affects patients (and even include them in governance conversations), and ensure that compliance with rules and regulations (e.g., HIPAA) is built into the evaluation of the model.

## How long should we evaluate?

AI models, trained on historical hospital data, inherently reflect existing hospital practices. However, these practices may evolve over time, potentially causing the model's performance to degrade. Thus, ensuring an AI model maintains performance over time remains challenging. Setting thresholds to identify significant shifts in real-world performance metrics is crucial. For example, drops in time savings, worsened patient outcomes, or sudden cost increases could signal issues with model effectiveness stemming from changes in practice, or sudden shifts in the patient population.  Continuous monitoring of these metrics via automated evaluations allows hospitals to keep an eye on any sudden changes. However, manual evaluations by human experts should also be conducted, weighing the time and effort required against the cost of human involvement. This way, machine learning practitioners can proactively address these shifts, updating models as needed to maintain their effectiveness. maries, understanding how model outputs interface with a wide variety of patients' experiences with the healthcare system may be important to consider as well. Equally important are policymakers and legal experts, who could help contextualize how this system affects patients (and even include them in governance conversations), and ensure that compliance with rules and regulations (e.g., HIPAA) is built into the evaluation of the model.

# Takeaways

- Standard metrics and benchmarks for evaluating machine learning models, while useful as a proxy for model performance, may not fully reflect the complex and nuanced nature of everyday clinical care delivery.

- For healthcare applications, human expert judgment can serve as a gold standard but is costly. Automated methods can be helpful for scaling, but need to be checked for their suitability within the clinical workflows.

- AI models reflect hospital practices in the data they are trained on and may break as hospital practices change over time. Continuous evaluation of model behavior over time can expose performance reductions, providing an opportunity to intervene. Sharing such model evaluation results with a combination of regulatory bodies, hospital leadership, patients, and the public is crucial to building trust in models.

- None of the guidance or recommendations provided here are exclusive to GenAI models, which are also vulnerable to discrepancies between benchmark evaluation and real-world usage. The surge in interest in GenAI reaffirms the importance of conscientious LLM evaluation centering on real-world priorities.

## Content Moderation

Social media companies process tens of millions of messages daily to identify content that needs to be moderated. At this scale, human moderators alone cannot review all content, making AI systems a critical component of the moderation workflow. These systems are most commonly used for flagging content that might be objectionable, automating initial reviews, and reducing the cognitive and emotional burdens on human moderators.
When successful, these systems not only reduce exposure to harmful content for both regular users and moderators but also improve trust in platform governance by generating explanations for enforcement decisions. Yet, when erroneous, these systems can wrongfully penalize legitimate speech, amplifying concerns about bias and fairness.

## What is being evaluated?

Traditionally automated content moderation has used a variety of benchmarks for identifying toxic language or other types of content that run afoul of the platform's policies. However, AI evaluation needs to consider more than just model performance: it needs to include estimates of the downstream effects of the system's decisions on the humans that interact with it. Beyond simply evaluating accuracy, AI evaluation must assess factors such as scalability, speed, the impact of false classifications, and even moderator well-being. A highly accurate model that operates too slowly is impractical in real-time moderation on commercial platforms, while a fast but imprecise model risks either overwhelming moderators with unnecessary flags or allowing harmful content to persist unchecked. False negatives—where harmful content is missed—can lead to real-world harm, including harassment, misinformation spread, or radicalization. Meanwhile, false positives—where innocuous content is mistakenly flagged—can disproportionately silence marginalized voices, inhibit open discourse, and erode user trust. Models can also be evaluated by their human impact: Exposing moderators to extremely toxic content is known to cause mental health issues, so models can be evaluated according to how infrequently humans (customers and moderators) are exposed. A comprehensive evaluation framework should not only quantify these risks but also weigh them against each other, ensuring that AI-assisted moderation aligns with broader platform policies, user expectations, and legal constraints.

## Who should evaluate?

The majority of GenAI evaluations assume there is some agreed-upon truth to compare models with. However, for content moderation, people may disagree—with good reason. While some content like explicit hate speech is clearly objectionable to most folks, other content may elicit different opinions. For example, some social groups targeted by offensive speech may view such content as much more offensive and needing to be moderated, while members of other groups may not perceive the content in the same way. Thus, for content moderation, evaluators need to come from different types of backgrounds to ensure that the values embodied by the AI system's judgments reflect community consensus and those harmed by the content have their voices included.

## How long should we evaluate?

People are endlessly creative, which, unfortunately, also extends into offensive communication. AI models trained for content moderation need to be regularly updated to reflect new forms of offensive language and any new changes to the platform's policies. As new models are developed, data leakage becomes a serious concern for content moderation; an AI model may seem to be very effective at identifying objectionable content but, in reality, the model may have been pretrained on the test data of an older benchmark.

## Takeaways

- Mistakes and correct actions by AI Models can have different implications on net benefit. Evaluation needs to be contextualized with the impact of the system on humans, both good and bad.

- AI Evaluations need to take into account the performance for different populations (e.g., demographic groups) and whether those populations' views are represented in the model's decisions.

# Summary of Recommendations

- **Domain practitioners** should design continuous and adaptive evaluation strategies that reflect real-world complexities. The evaluations should also ensure the models are tested not only for performance but also for fairness, reliability, efficiency, and usability in practical settings. It is crucial to incorporate human expertise and domain-specific knowledge into the evaluation process and to use an understanding of how AI models interact with existing workflows.

- **Policymakers** should focus on regulating the outcomes of AI systems rather than specific technologies. This approach allows policies to remain flexible and adaptive as AI evolves, while ensuring that key outcomes such as fairness, transparency, environmental impact, and societal impact are prioritized. Policymakers should also ensure that evaluation processes are transparent and promote trust and accountability among stakeholders.

- **Business leaders** should also invest in dynamic, outcome-oriented evaluation frameworks that go beyond traditional metrics to assess the broader impact of AI systems on users, operations, and society. It's important to foster collaboration between technical teams and domain experts. This would allow the AI deployment to align with long-term business goals and ethical standards and provide transparency to stakeholders.

- **Future evaluation designers** should develop multi-faceted frameworks that combine automated metrics with human judgment to assess AI models in both controlled and real-world scenarios. The evaluation methods should be adaptable to allow continuous monitoring and updates as AI systems evolve and should consider ethical, social, and domain-specific factors.

- **Sponsors of funding agencies** should prioritize investments in efforts and innovations to develop comprehensive, outcome-oriented evaluation frameworks that assess AI systems' real-world performance, fairness, and societal impact. Resources and funding should be allocated not only for model development but also for continuous evaluation and adaptation, to ensure AI systems remain reliable. Additionally, interdisciplinary approaches that bring together technical, ethical, and domain-specific expertise should be supported to create robust evaluation methods.

# Acknowledgments

This white paper was primarily authored by members of the AI Lab at the University of Michigan and its partners on the AI Lab Advisory Board, along with collaborators from related disciplines. The content reflects only the authors' views and does not represent those of their respective institutions.

We extend our special thanks to Christina Certo for her invaluable assistance in organizing and coordinating meetings and to Aurelia Bunescu for her support with graphic design.